\UseRawInputEncoding
\documentclass[letterpaper, 10 pt, conference]{ieeeconf}  

\IEEEoverridecommandlockouts       

\usepackage{multirow}  
\usepackage{arydshln}  
\usepackage{graphicx}
\usepackage{epstopdf}
\usepackage{threeparttable}  
\usepackage{placeins}
\usepackage{xcolor}
\usepackage{amsmath}
\definecolor{darkred}{rgb}{0.55, 0, 0} 

\usepackage[
  breaklinks=true,
  colorlinks,
  linkcolor=darkred,
  urlcolor=magenta,
  citecolor=teal,
  bookmarks=false
]{hyperref}
\usepackage{booktabs} 

\title{\LARGE \bf
Articulated Object Manipulation using Online Axis Estimation with SAM2-Based Tracking
}

\author{Xi Wang$^{\ast,2}$, Tianxing Chen$^{\ast,1,2}$, Qiaojun Yu$^{\ast, 3}$, \\Tianling Xu$^{4}$, Zanxin Chen$^{2}$, Yiting Fu$^{2}$, Ziqi He$^{2}$, Cewu Lu$^{3,\dag}$, Yao Mu$^{1,\dag}$ and Ping Luo$^{1,\dag}$ \\ $^{*}$ Equal Contribution, $^{\dag}$ Corresponding Author\\
\href{https://hytidel.github.io/video-tracking-for-axis-estimation/}{https://hytidel.github.io/video-tracking-for-axis-estimation/}
\thanks{$^{1}$Yao Mu and Ping Luo are with Department of Computer Science, the University of Hong Kong, Hong Kong, China. Tianxing Chen did this work during the University of Hong Kong internship.}%
\thanks{$^{2}$Xi Wang, Tianxing Chen, Zanxin Chen, Yiting Fu and Ziqi He are with College of Computer Science and Software Engineering, Shenzhen University, Shenzhen, China.}%
\thanks{$^{3}$Qiaojun Yu and $\dag$Cewu Lu are with Shanghai Jiaotong University, Shanghai, China.}
\thanks{$^{4}$Tianling Xu is with the Southern University of Science and Technology.}
}

\begin{document}


\maketitle
\thispagestyle{empty}
\pagestyle{empty}

\begin{abstract}
Articulated object manipulation requires precise modeling, where understanding the 3D motion constraints of individual articulated components is crucial. Prior research has leveraged interactive perception to facilitate manipulation; however, open-loop approaches often fail to account for the interaction dynamics, limiting their effectiveness. To overcome this limitation, we propose a closed-loop pipeline that integrates interactive perception with online axis estimation and optimization. Our method processes sequentially segmented 3D point clouds of manipulated object parts, continuously refining the motion model throughout the manipulation process.
Specifically, our approach builds upon any interactive perception technique, inducing slight object movements to capture point cloud frames of the evolving dynamic scene. These point clouds are segmented using the Segment Anything Model 2 (SAM2) [Ravi et al.], where the moving object part is accurately masked for online axis estimation. The estimated axis then guides subsequent robotic motion planning, ensuring precise and adaptive manipulation.
Experimental results in simulated environments demonstrate that our method outperforms baseline approaches, particularly in tasks requiring precise axis-based control. By integrating real-time perception with online optimization, our framework significantly improves both the accuracy and efficiency of articulated object manipulation.
\end{abstract}
\section{INTRODUCTION}

Robotic manipulation has a wide range of applications, including industrial automation~\cite{afrin2024ai}, medical surgery~\cite{picozzi2024advances}, and warehouse logistics~\cite{sodiya2024ai}. Among various manipulation tasks, those involving articulated objects, such as doors and drawers, pose significant challenges~\cite{xiong2024adaptive}. This complexity arises from the need to understand not only the overall geometry of the object but also the composition and kinematic relationships between its articulated components~\cite{jiang2022ditto}.

Traditional manipulation techniques often rely on predefined kinematic models~\cite{gou2021rgbmatters, gao2021kpamsc} or open-loop control~\cite{wang2020learning, wu2022vatmart}. However, these approaches struggle to dynamically adapt to real-world interactions, particularly in the presence of uncertainties. A key limitation of open-loop methods is the absence of feedback regulation, preventing the robot from 

\begin{figure}[thbp]
    \centering
    \includegraphics[width=0.44\textwidth]{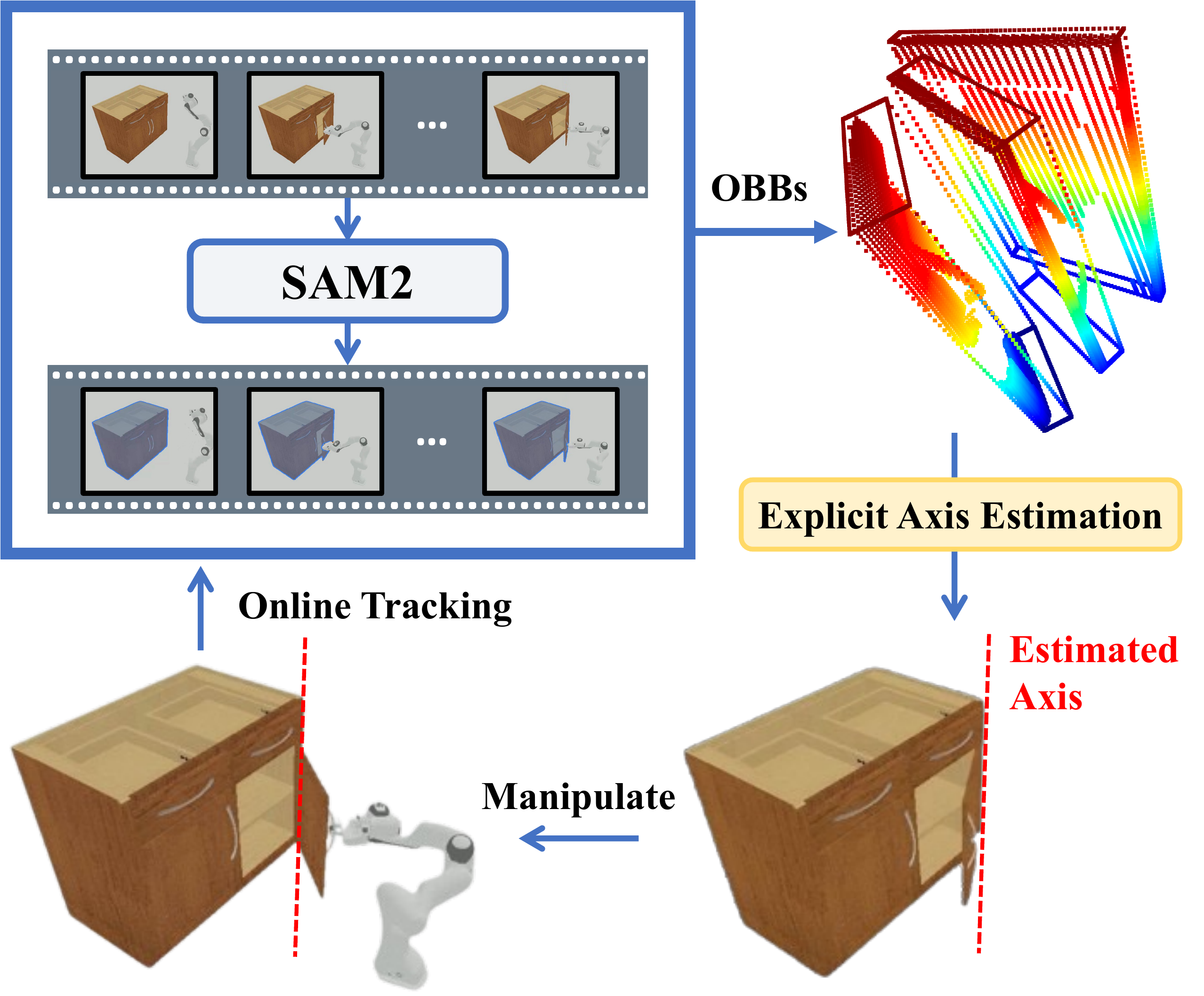}
    \caption{
        We introduces a closed-loop pipeline integrating SAM2 to track the articulated object throughout the manipulation process captured by an RGB-D camera. 
        The masks are subsequently used to segment out the point cloud of the articulated object, followed by explicitly estimating the joint axis using the oriented bounding boxes (OBBs) of the motion part derived from the point cloud. 
        The robot's action is instructed under the guidance of the motion axis to perform further manipulation for the next loop. 
    }
    \label{fig-teaser}
    \vspace{-2pt}
\end{figure}

\noindent adjusting its actions based on real-time interaction dynamics. Without continuous feedback, the system fails to accommodate changes in the object's state, resulting in inefficiencies and inaccuracies in manipulation tasks~\cite{wang2025adamanip}.


Recently, interactive perception has emerged as a promising approach to address these challenges \cite{jiang2022ditto, wu2022vatmart, hau2015artiperception, an2024rgbmanip}. 
Interactive perception aims to bridge the gap between perception and action by using sensory feedback from these interactions. 
By actively interacting with the object, robots can gather real-time sensory data that provides insights into the structure and kinematics of objects. 
While these methods effectively provide information about the object’s state, it lacks a critical component — the evolving dynamic interactions between the robot and the articulated object over time. 
This omission limits the robot’s ability to adapt its manipulation policy in real time as the object’s state changes, leading to inefficiencies in tasks that require precise control.

To address this limitation, in this paper, we propose a novel closed-loop pipeline that integrates interactive perception with online refined axis estimation, providing the robot with helpful guidance for axis-aware manipulation. 
The basic idea is illustrated in Fig. \ref{fig-teaser}. 
Our approach builds on interactive perception, and enhances it under the guidance of an online refined motion axis, which is derived from segmented 3D point clouds.
Specifically, we leverage any interactive perception technique, such as RGBManip~\cite{an2024rgbmanip}, as a foundation method to induce slight object movements and generate dynamic 3D point cloud frames. 
These frames are then processed using an advanced segmentation pipeline, with Grounding DINO~\cite{liu2023grounding} as object detector and Segment Anything Model 2 (SAM2)~\cite{ravi2024sam2} for object segmentation, which identifies and isolates the point cloud of articulated object. 
By masking out the moving components of the object, we can explicitly calculate the motion axis with the oriented bounding boxes (OBBs), which in turn informs the robot’s subsequent manipulation action.
In contrast to previous studies, our closed-loop integration of interactive perception with online refined axis estimation allows for more precise and adaptive control, addressing the limitations of traditional methods that operate in an open-loop fashion. 
By continuously updating the robot’s understanding of the object’s kinematic state, our approach ensures that the robot maintains effective control throughout the manipulation task.
We validate our approach through tasks that require precise axis-based control, such as drawer-opening and door-opening. 
Simulated results demonstrate that our method significantly outperforms pure interactive perception techniques, enhancing the precision of manipulation tasks involving articulated objects, and shows strong generalization across different articulated objects. 

To summarize, the three key contributions of our paper are as follows:

\begin{enumerate}
    \item \textbf{Utilization of 2D foundation models for 3D-based manipulation. } 
    Our pipeline integrates advanced 2D vision techniques, including Grounding DINO for object detection and SAM2 for video segmentation and tracking, and extends their capabilities to 3D point cloud segmentation and tracking. This enables object-level iterative dynamics modeling of articulated objects, allowing for precise and continuous tracking of their motion throughout the manipulation process.

    \item \textbf{An online motion axis estimation method based on part-level iterative dynamics. }
    We derive the moving components of the articulated objects from the dynamic evolution of the object point cloud, and explicitly calculate the motion axis in an online manner.  

    \item \textbf{Closed-loop integration of interactive perception and online axis estimation. } 
    By continuously refining the motion axis estimation based on the updates of 3D point cloud, our approach improves the manipulation precision, and enables adaptive and precise control throughout the manipulation process, overcoming the limitations of open-loop methods.

    
\end{enumerate}

\section{RELATED WORK}

\subsection{Vision-based Robotic Manipulation}


Visual perception is crucial for robotic manipulation, given its heavy reliance on visual modalities. 
Existing works adopt various inputs, with each modality exhibiting distinct advantages and limitations. 
While RGB images offer rich texture, they lack depth information crucial for grasping~\cite{gou2021rgbmatters}. 
To address this, some methods combine RGB with depth maps~\cite{Xu_2022UMPNet,wu2020graspproposalnetworksendtoend}, though RGB-D data often suffers from noise on transparent or reflective surfaces~\cite{sajjan2019cleargrasp3dshapeestimation}.

Unlike approaches that switch across modalities~\cite{ichnowski2021dexnerfusingneuralradiance,dai2023graspnerfmultiviewbased6dofgrasp,zheng2024gaussiangrasper3dlanguagegaussian}, RGBManip~\cite{an2024rgbmanip} uses only multi-view RGB to estimate 6D object poses. However, its single end-effector camera can miss key interaction dynamics. 
Point-cloud-based methods, such as Where2Act~\cite{mo2021where2actpixelsactionsarticulated} and SAGCI-System~\cite{lv2022sagcisystemsampleefficientgeneralizablecompositional}, excel at articulating action affordances, with RLAfford~\cite{wu2020graspproposalnetworksendtoend} emphasizing interaction dynamics. Flowbot3D~\cite{eisner2024flowbot3dlearning3darticulation} leverages motion flows to handle dynamic articulated objects.

Our approach strives to integrate the strengths of RGB-based methods and point-cloud-based techniques by tracking RGB inputs and segmenting the point clouds corresponding to the motion part of articulated objects. 
In order to address the limitation of open-loop methods, we integrate an additional RGB-D camera into the scene, specifically aimed at capturing the manipulation process.

\begin{figure*}[htbp]
\centering
\includegraphics[width=1.0\textwidth]{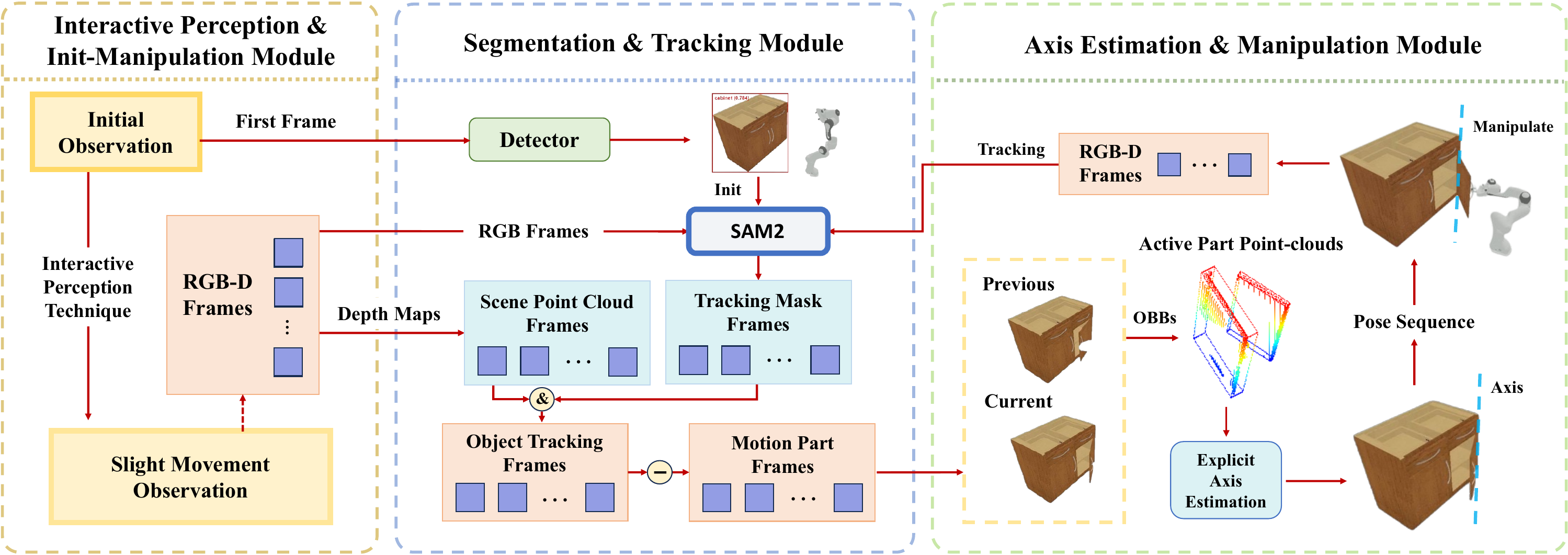}
\caption{In our pipeline, an RGB-D camera captures the dynamic scene, which is induced by the slight movement from the Interactive Perception \& Init-Manipulation Module. The captured scene is then processed by the Tracking \& Segmentation Module, which tracks and segments the moving part of the articulated object at a 3D level. This segmented data is subsequently passed to the Axis Estimation \& Manipulation Module. Here, the motion axis is explicitly calculated, providing informed guidance for the robot's manipulation policy.}
\label{fig-pipeline}
\end{figure*}

\subsection{3D Point Cloud Segmentation Empowered by 2D Foundation Models}

Segmentation techniques for 2D images have become highly sophisticated, with the Segment Anything Model (SAM)~\cite{kirillov2023segment} as a prominent example. 
However, transferring these methods to 3D point clouds remains challenging, limiting their applicability in robotic manipulation.

DeepLabv3~\cite{chen2017rethinkingatrousconvolutionsemantic} adapts 2D segmentation to 3D by projecting point clouds onto 2D planes~\cite{chen2024g3flow}, which is robust at scale but loses accuracy due to projection. 
To address this, SAM-based 3D approaches have emerged. 
For instance, Point-SAM~\cite{zhou2024pointsampromptable3dsegmentation} extends SAM to generate segmentation masks from point or mask prompts, while PartSLIP~\cite{liu2023partsliplowshotsegmentation3d} and PartSLIP++~\cite{zhou2023partslipenhancinglowshot3d} leverage GLIP~\cite{li2022groundedlanguageimagepretraining} and SAM for fine-grained part-level segmentation.

In dynamic scenes, video segmentation models such as SAM2 preserve motion cues more effectively. 
Our method builds on the RGBManip environment by integrating a single RGB-D camera and applying robust noise-handling. 
By masking the moving parts of the target objects in the point cloud, we capture motion information that better informs the manipulation policy.

\subsection{Axis Estimation for Articulated Objects}

Axis estimation plays a critical role in the perception, comprehension and manipulation of articulated objects. 
Extensive research contributes various techniques to tackle the complexities of recognizing and estimating the motion constraints and joint types of articulated objects. 

Individual observations are often inherently ambiguous for articulated objects. \cite{Bohg2017interactiveperceptionleveragingactioninPerceptionandPerceptioninAction}~highlighted the important role of interactive perception in manipulation. 
Subsequent research has extensively adopted interactive perception techniques to effectively mitigate the ambiguity inherent in single observations. 
Act the Part~\cite{gadre2021actpartlearninginteraction} enabled robots to learn how to interact with articulated objects, which contributes to discovering and segmenting their parts. 
\cite{Martin2014onlineinteractiveperceptionofarticulatedobjects}~presented an online interactive perception system based on task-specific priors to extract kinematic and dynamic models of articulated objects. 

However, previous studies on axis estimation are typically open-loop, which causes a loss of the interaction dynamics. 
To address this issue, \cite{karayiannidis2016adaptive}~utilizing force/torque sensor measurements to estimate the motion direction and the orientation of the axis, inferring the type of joint without prior knowledge of the object's kinematics. 
It employs closed-loop control relies on a velocity controller, which ensures that the end-effector moves at the desired velocity. 

Our approach also employs closed-loop axis estimation. 
Unlike~\cite{karayiannidis2016adaptive} which adopts the implicit estimation via kinematic models, we leverage the geometric properties of motion to explicitly compute the axis of the articulated object.



\section{METHOD}

We introduce a novel approach for manipulating articulated objects, guided by explicit axis estimation derived from SAM2-based tracking. 
As depicted in Fig.~\ref{fig-pipeline}, our pipeline consists of 3 core modules: (1) Interactive Perception \& Init-Manipulation Module, (2) Segmentation \& Tracking Module and (3) Axis Estimation \& Manipulation Module.

Initially, the robot applies any interactive perception methods to manipulate the articulated object, creating a slight displacement. 
Our method leverages the robust tracking capabilities of SAM2 to capture the 3D dynamics of articulated objects. 
After point cloud augmentation, we can obtain the Oriented Bounding Boxes (OBBs) and explicitly compute the precise axis information of the articulated object to drive the robotic arm for manipulation.

\subsection{Interactive Perception \& Init-Manipulation Module}

The Interactive Perception \& Init-Manipulation Module employs any interactive perception methods to grasp the handle and perform initiate manipulation to articulated objects, producing slight displacements for further analysis. 
Our method incorporates RGBManip as the default implementation for this module, chosen for its recent success in RGB-only interactive perception. 
It is worth noting that RGBManip employs the original SAM-based multi-view object pose estimation for detecting and grasping the handle. 

As RGBManip initiates the manipulation of articulated objects, an additional camera periodically records dynamic RGB-D data. 
The captured RGB data will be leveraged for subsequent segmentation and tracking tasks, while the corresponding depth data facilitate the reconstruction of the 3D point cloud of the evolving dynamic scene. 

\subsection{Segmentation \& Tracking Module} 
We exploit Grounding DINO~\cite{liu2023grounding} on the initial RGB frame to generate an anchor box for the target object using text prompt, such as ``cabinet".
The first RGB frame serves as the starting point for SAM2's tracking process~\cite{ravi2024sam2}. 
The anchor box, corresponding to the target object, is input into SAM2 as a box prompt, enabling it to continuously track the object, and provide a mask for it throughout each moment of the dynamic manipulation process. 
With the object's mask, we can extract the corresponding portion of the 3D point cloud representing the object from the dynamic scene. 

We filter the raw object point cloud $P = \{p_i\}_{i = 1}^n$ obtained above to remove outliers, ensuring that the filtered point cloud represents the region of interest of articulated object. 
Specifically, a point $p \in P$ is retained if and only if 
\begin{equation}
    |U^\circ (p, r)| \geq \epsilon, 
\end{equation}
in which 
\begin{equation}
    U^\circ (p, r) = \{q \mid q \in P, 0 < |p - q| \leq r\}, 
\end{equation}
and $|\cdot|$ is the number of element in the set. 

We then compute the oriented bounding box (OBB) for the entire articulated object on the initial frame. 
Subsequently, by subtracting this OBB from each frame's point cloud and applying another round of noise reduction, we obtain the point cloud representing the moving parts of the articulated objects. 
For implementation details, the segmentation can be refined by removing the protruding handle from the cabinet's OBB to obtain a tighter OBB of the cabinet's body, leading to a more precise motion-part segmentation. 

\subsection{Axis Estimation \& Manipulation Module}

The Axis Estimation \& Manipulation Module calculates the axis of motion based on the segmented point cloud representing the moving parts of the articulated object identified by the Segmentation \& Tracking Module. 
Task-specific geometric priors play an essential role in achieving accurate axis estimation, i.e., for a prismatic joint, the moving parts translate along the axis, whereas for a revolute joint, the moving parts rotate around the axis. 
By leveraging these task-based priors, we are able to explicitly calculate the joint's axis. 

Specifically, consider a single action of robotic manipulation, where the OBBs of the initial and final point cloud of the motion components of the articulated object are denoted as $obb_{\text{st}}$ and $obb_{\text{ed}}$, with their centers designated as $O_{\text{st}}$ and $O_{\text{ed}}$ respectively.

\begin{itemize}
    \item \textbf{Prismatic joint. } 
    The axis pivot is defined as $O_{\text{st}}$, and the axis direction is estimated as the direction $\overrightarrow{d} = \overrightarrow{O_{\text{st}}O_{\text{ed}}}$.

    \item \textbf{Revolute joint. }
    The axis pivot point is derived by identifying the intersection of the mid-perpendiculars along the longer edges in the top-down view of $obb_{\text{st}}$ and $obb_{\text{ed}}$. 
    With the axis point $P$ established, the axis direction is ascertained by evaluating the sign of the dot product $\overrightarrow{d} \cdot \overrightarrow{t}$, where $\overrightarrow{t} = \overrightarrow{O_{\text{st}} P} \times \overrightarrow{z}$, among which $\overrightarrow{z}$ represents the positive direction of the z-axis.
\end{itemize}

We propose a closed-loop axis estimation refinement method predicated on a straightforward observation: as the manipulation of the joint assembly increases, the point cloud of its moving component becomes progressively more amenable to accurate reconstruction. 
The enhancement in the fidelity of the point cloud data, in turn, furnishes the axis estimation process with inputs that are increasingly precise and reliable. 

For implementation, after each manipulation, we invoke the Segmentation \& Tracking Module and the Axis Estimation Module to ascertain the current axis estimation, in which the window of frame indices $[\text{st}, \text{ed}]$ is progressively shifted as the process unfolds, while maintaining an appropriate length of the interval $[\text{st}, \text{ed}]$.
The Manipulation Module then guides the robot's subsequent actions according to this estimation. 
The interactive process reiterates until the robot has executed all designated actions, with the axis estimation being continuously refined throughout the procedure.

\section{EXPERIMENT}


\begin{table*}[htb]
\centering
\caption{Quantitative comparison between our method and baselines}
\label{table-comparison_with_baselines}
\begin{tabular}{c | c |c c|c c|c c|c c}
\hline
\multirow{2}{*}{\textbf{Methods}} & \multirow{2}{*}{\textbf{Modality}} & \multicolumn{2}{c|}{\textbf{Open Door 8.6°}} & \multicolumn{2}{c|}{\textbf{Open Door 45°}} & \multicolumn{2}{c|}{\textbf{Open Drawer 15 cm}} & \multicolumn{2}{c}{\textbf{Open Drawer 30cm}} \\
& & Train & Test & Train & Test & Train & Test & Train & Test \\ \hline

DrQ-v2 
\footnotemark[1] & RGB
    & 1.8 & 2.5 & 0.8 & 0.8 & 1.9 & 1.0 & 1.4 & 0.5  \\
LookCloser 
\footnotemark[1] & RGB
    & 1.5 & 1.25 & 0.8 & 0.8 & 0.8 & 0.0 & 0.0 & 0.0 \\
 RGBManip 
 \footnotemark[2] & RGB
    & 75.0 & 82.0 & 47.0 & 47.0 & 56.0 & 64.0 & 46.0 & 45.0 \\ 

\hdashline

Where2Act 
\footnotemark[1] & PCD
    & 8.0 & 7.0 & 1.8 & 2.0 & 5.9 & 7.5 & 1.1 & 0.6 \\
Flowbot3D \footnotemark[1] & PCD
    & 19.5 & 20.4 & 6.8 & 6.4 & 27.3 & 25.8 & 16.9 & 11.3 \\
UMPNet \footnotemark[1] & PCD
    & 27.1 & 28.1 & 11.0 & 10.9 & 16.6 & 18.8 & 4.4 & 5.6 \\
GAPartNet \footnotemark[1] & PCD
    & 69.5 & 74.5 & 39.4 & 43.6 & 50.6 & 59.3 & 44.6 & 48.6 \\ 

\hdashline

\textbf{Ours} & RGB + PCD
    & \textbf{87.0} & \textbf{88.0} & \textbf{54.0} & \textbf{54.0} & \textbf{68.0} & \textbf{85.0} & \textbf{59.0} & \textbf{68.0} \\\hline
\end{tabular}
\end{table*}


\begin{table*}[htb]
\centering
\caption{More challenging tasks for opening door}
\label{table-more_for_opening_door}
\begin{tabular}{c|c c|c c|c c|c c}
\hline
\multirow{2}{*}{\textbf{Methods}} & \multicolumn{2}{c|}{\textbf{20°}} & \multicolumn{2}{c|}{\textbf{30°}} & \multicolumn{2}{c|}{\textbf{40°}} & \multicolumn{2}{c}{\textbf{50°}} \\
  & Train & Test & Train & Test & Train & Test & Train & Test \\ \hline
RGBManip & 72.0 & 75.0 & 66.0 & 66.0 & 53.0 & 56.0 & 39.0 & 38.0 \\
\textbf{Ours} & \textbf{75.0} & \textbf{77.0} & \textbf{72.0} & \textbf{70.0} & \textbf{62.0} & \textbf{59.0} & \textbf{44.0} & \textbf{44.0} \\ \hline
\end{tabular}

\vspace{2.5pt}

\begin{tabular}{c|c c|c c|c c|c c}
\hline
\multirow{2}{*}{\textbf{Methods}} & \multicolumn{2}{c|}{\textbf{55°}} & \multicolumn{2}{c|}{\textbf{60°}} & \multicolumn{2}{c|}{\textbf{65°}} & \multicolumn{2}{c}{\textbf{70°}} \\
 & Train & Test & Train & Test & Train & Test & Train & Test \\ \hline
RGBManip & 32.0 & 32.0 & 26.0 & 28.0 & \textbf{23.0} & 16.0 & \textbf{22.0} & 13.0 \\
\textbf{Ours} & \textbf{38.0} & \textbf{41.0} & \textbf{28.0} & \textbf{35.0} & 22.0 & \textbf{27.0} & 19.0 & \textbf{22.0} \\ \hline
\end{tabular}

\end{table*}
\begin{table*}[htb]
\centering
\caption{More challenging tasks for opening drawer}
\label{table-more_for_opening_drawer}
\begin{tabular}{c|c c|c c|c c|c c|c c}
\hline
\multirow{2}{*}{\textbf{Methods}} & \multicolumn{2}{c|}{\textbf{20 cm}} & \multicolumn{2}{c|}{\textbf{25 cm}} & \multicolumn{2}{c|}{\textbf{35 cm}} & \multicolumn{2}{c|}{\textbf{40 cm}} & \multicolumn{2}{c}{\textbf{45 cm}}  \\
 & Train & Test & Train & Test & Train & Test & Train & Test & Train & Test \\ \hline
 
RGBManip  & 51.0 & 61.0 & 48.0 & 57.0 & 41.0 & 43.0 & 34.0 & 31.0 & 24.0 & 14.0  \\
\textbf{Ours} & \textbf{64.0} & \textbf{84.0} & \textbf{61.0} & \textbf{76.0} & \textbf{57.0} & \textbf{63.0} & \textbf{50.0} & \textbf{58.0} & \textbf{35.0} & \textbf{41.0} \\ \hline
\end{tabular}

\vspace{2.5pt}

\end{table*}

\subsection{Task Settings} \label{sec-task_settings}

In this study, we specifically focus on 1-DoF doors and drawers for experimental validation.
We evaluate the performance in scenarios with large ranges of door-opening and drawer-extension.

In each task, the articulated objects, placed with limited random position and rotation, begins in a closed state, and the robotic arm is initially positioned randomly in front of it. The arm must accomplish the designated manipulation goal, either opening the drawer or door to a specific degree or range.

The task settings are detailed as follows: 

\begin{itemize}
    \item \textbf{Open Door:} The agent needs to open the door larger than 8.6°, 10°, 20°, 30°, 40°, 45°, 50°, 55°, 60°, 65°, 70°. 
    \item \textbf{Open Drawer:} The agent needs to open the drawer larger than 10 cm, 15 cm, 20 cm, 25 cm, 30 cm, 35 cm, 40 cm, 45 cm.
\end{itemize}

To ensure fairness in comparison, all RGBManip components employ task-specific \textit{adapose} as the pose estimator and \textit{heuristic pose} as the controller among different tasks. 
All methods are compared under equivalent total step sizes, though different methods may allocate step sizes differently based on their respective policies. 

We select $r = 0.05$ and $\epsilon = 100$ as the hyperparameters for point cloud augmentation. 

For each task, we evaluate our methods compared with RGBManip and other baselines separately on RGBManip's training set and testing set. 
Success rates of the first 100 experiments are used as metrics for comparison respectively. 


\subsection{Baselines}

We benchmark our approach against seven existing techniques (including RGBManip). 
Below is a brief summary of each comparison method. 


\begin{itemize}
    \item \textbf{DrQ-v2 \cite{yarats2021masteringvisualcontinuouscontrol}:} Based on reinforcement learning (RL), it takes in the robot's state and RGB image to determine the desired 6D pose of the robot's end-effector. 
    \item \textbf{LookCloser \cite{jangir2022lookcloserbridgingegocentric}:} A multi-perspective RL model that leverages multi-view inputs and visual transformers to amalgamate data from various angles. 
    \item \textbf{RGBManip \cite{an2024rgbmanip}:} Utilizes RGB-only visual input to directly estimates the 6D pose of objects from multi-view images. 
    \item \textbf{Where2Act \cite{mo2021where2actpixelsactionsarticulated}:} Processes point-cloud data to estimate the best point of interaction for manipulation. 
    \item \textbf{Flowbot3D \cite{eisner2024flowbot3dlearning3darticulation}:} Predicts point-wise motion, or ``flow", within the point cloud. The point with the highest motion magnitude is selected for interaction. 
    \item \textbf{UMPNet \cite{Xu_2022UMPNet}:} Utilizing RGB-D images, it predicts an action point in the image and projects it into 3D space using depth data. 
    \item \textbf{GAPartNet \cite{geng2023gapartnetcrosscategorydomaingeneralizableobject}:} A pose-centric method which predicts the pose of an object's part from point-cloud data. 
\end{itemize}

\subsection{Quantitative Results}

Quantitative results of basic tasks are summarized in Table. \ref{table-comparison_with_baselines}, where both our method and RGBManip almost outperform other baseline approaches while Ours consistently surpasses RGBManip in basic tasks. 

\begin{itemize}
    \item Granular results for more challenging door-opening tasks are presented in Table. \ref{table-more_for_opening_door}, with the corresponding line chart showing success rates with regard to door-opening angles depicted in Fig. \ref{fig-more_for_opening_door}. 
    
    \item Fine-grained results for more challenging drawer-opening tasks are provided in Table. \ref{table-more_for_opening_drawer}, with the corresponding line chart illustrating success rates at various drawer-opening distances shown in Fig. \ref{fig-more_for_opening_drawer}. 
\end{itemize}

Experimental results show that Ours consistently outperforms RGBManip with a significant enhancement in success rates, showing robustness across various types of cabinets. 

\footnotetext[1]{Experimental results derived from TABLE I of the paper \textit{RGBManip: Monocular Image-based Robotic Manipulation through Active Object Pose Estimation}
\cite{an2024rgbmanip}.}
\footnotetext[2]{Results reproduced using RGBManip under the settings outlined in Sec. \ref{sec-task_settings}.}

\subsection{Real-World Deployment}

\begin{figure*}[htbp]
\centering
\includegraphics[width=1.0\textwidth]{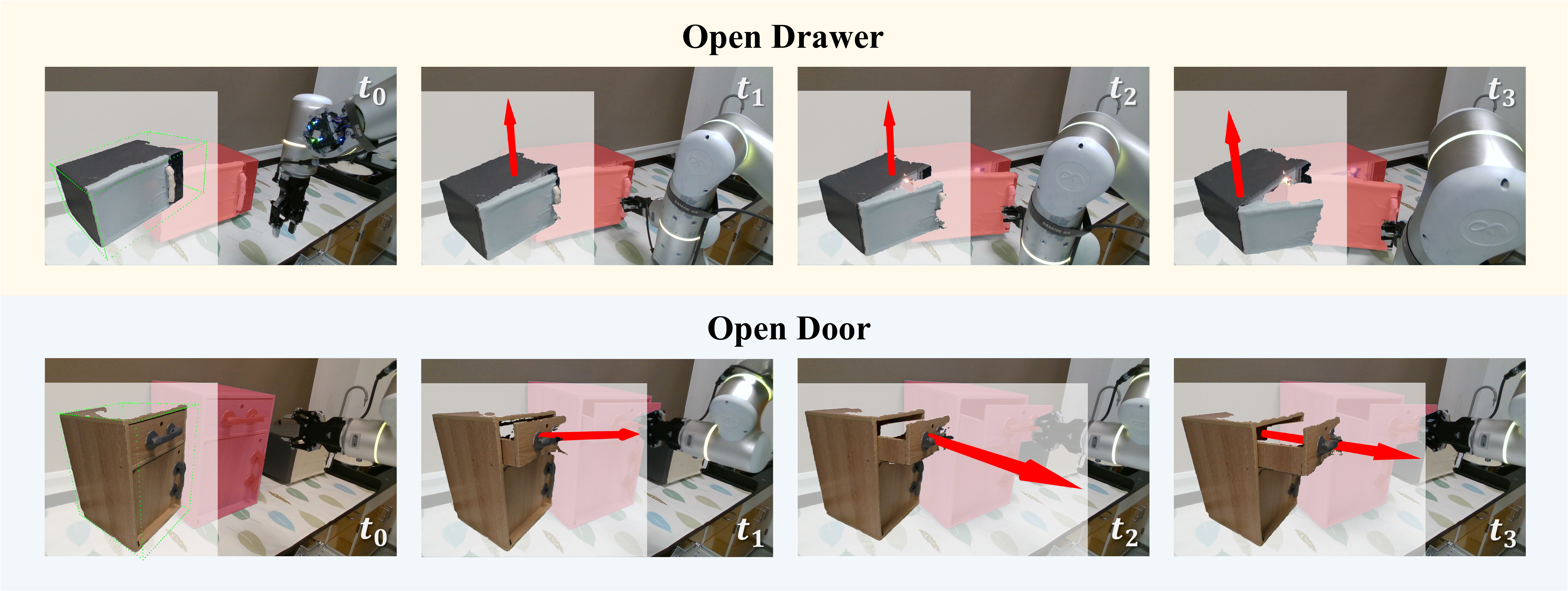}
\caption{ 
    \textbf{Visualization of axis estimation for real-world ``Open Drawer'' and ``Open Door'' tasks. }
    The initial moment and the three manipulation moments are shown, with visualization of the RGB tracking obtained from SAM2 (background), the reconstructed point cloud of the target object (bottom-left corner), the OBBs (green dashed-line boxes at $t_0$) and the axis (red arrows) estimated with our method. 
}
\label{fig:real-robot}
\end{figure*}

To validate the effectiveness of our method in real-world deployment, we demonstrate the complete axis estimation process for the door-opening and drawer-opening tasks. 

We employ the D415 depth camera. 
The hyperparameters are set to $r = 1.3$ and $\epsilon = 1$, due to the sparsity of the point cloud reconstructed from the real-world depth map. 

Empirical results are shown in Fig.~\ref{fig:real-robot}. 
Despite the presence of noise in real-world point clouds, our method remains robust. 
This is attributes to our point cloud augmentation method, which enables robust computation of OBBs, leading to accurate and reliable axis estimation in noisy scenarios. 

\section{Analysis} \label{analysis} 

The analysis section delves into the specific experimental results to elucidate the practical advantages of our online axis estimation approach in the context of articulated object manipulation. Our method's performance is grounded in the empirical data obtained from the experiments, which are analyzed below to provide a detailed understanding of the improvements achieved.

\subsection{Online Axis Estimation vs. Traditional Methods}

\begin{figure}[htbp]
\centering
\includegraphics[scale=0.2]{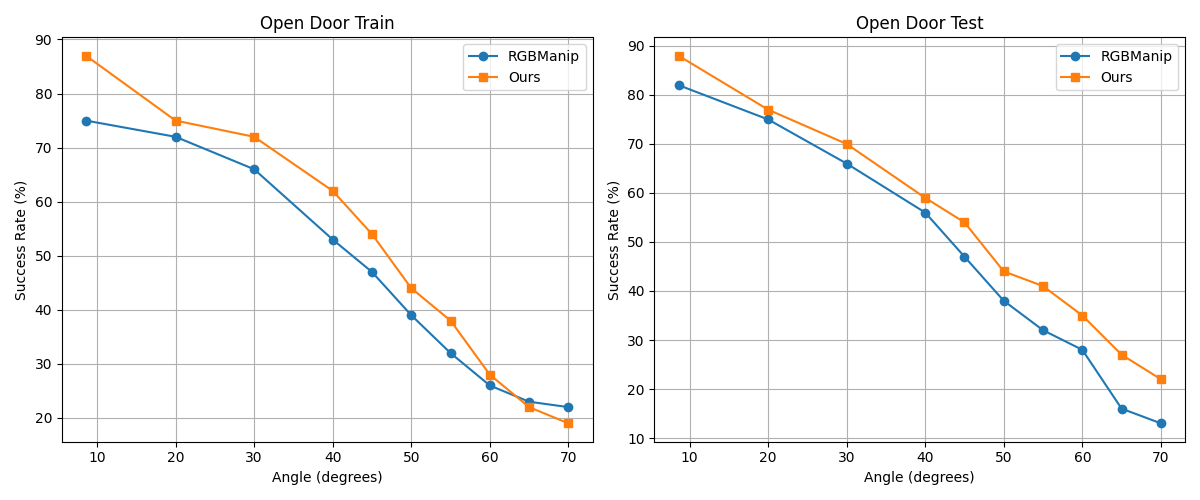}
\caption{Success rate of more challenging tasks for opening door. }
\label{fig-more_for_opening_door}
\end{figure}

\begin{figure}[htbp]
\centering
\includegraphics[scale=0.2]{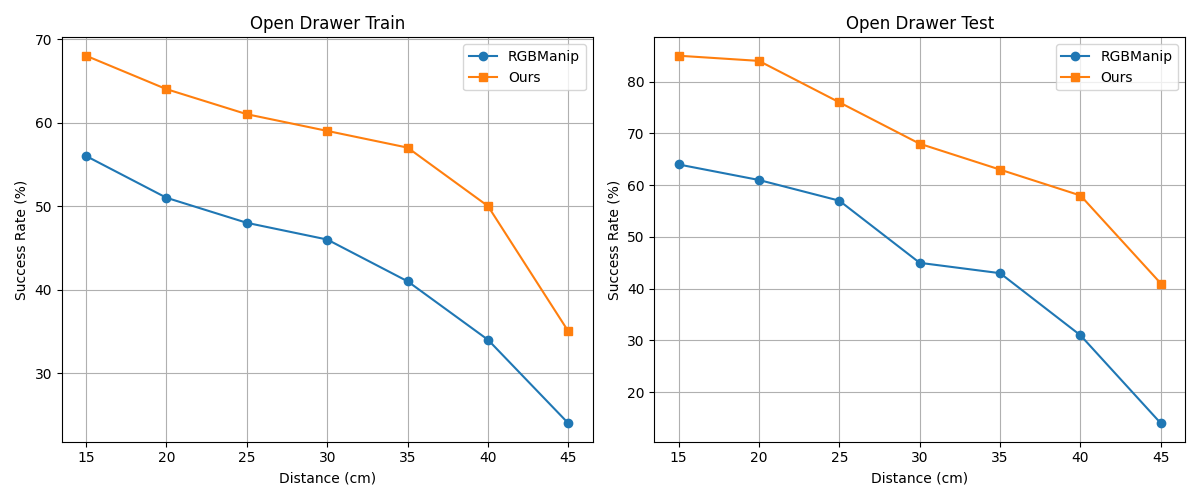}
\caption{Success rate of more challenging tasks for opening drawer. }
\label{fig-more_for_opening_drawer}
\end{figure}
Our experiments clearly demonstrate the superiority of our online axis estimation approach over traditional methods, especially in tasks that demand precise control such as door and drawer opening. As illustrated in Table \ref{table-comparison_with_baselines}, our method achieves an impressive 87.0\% success rate in the training set and 88.0\% in the test set for the ``Open Door" task, significantly outperforming RGBManip, which records 75.0\% and 82.0\% respectively. This notable enhancement stems directly from the continuous refinement of the axis estimation, empowering the robot to adapt its actions in response to the most current interaction dynamics.

Furthermore, in the more demanding versions of these tasks, depicted in Figures \ref{fig-more_for_opening_door} and \ref{fig-more_for_opening_drawer}, our method's reliance on ongoing online axis estimation reveals enhanced robustness in handling large-amplitude movements of articulated objects. The experimental data consistently show that our method sustains higher success rates as the complexity of tasks escalates, whether it involves opening doors to wider angles or drawers to greater extents. The online axis estimation process is pivotal in this regard, enabling our system to dynamically adjust to real-time interaction feedback and ensuring precise control over articulated objects. The continuous axis refinement, aligned with the object's changing state, is essential for the manipulation's accuracy and efficiency, particularly in tasks with substantial state variations. The real-time feedback loop is crucial, enabling the robot to make swift and accurate operational adjustments that accommodate the subtleties of the interaction, thereby achieving a higher success rate in task completion. This robust performance underscores the strength of online axis estimation in providing reliable and responsive control in sophisticated robotic manipulation scenarios.

\subsection{Consistency Across Various Manipulation Scenarios}
Our method's consistent performance across a range of manipulation scenarios highlights its robustness and versatility. The success rates in both door and drawer opening tasks, as depicted in Tables \ref{table-more_for_opening_door} and \ref{table-more_for_opening_drawer}, consistently show higher rates for our method, indicating that the online axis estimation is effective regardless of the specific manipulation task. This consistency is a significant advantage over methods that may perform well in one scenario but falter in others.

\section{CONCLUSION}

In this research, we introduced a novel approach to articulated object manipulation that employs online axis estimation integrated with SAM2-based tracking. We utilizes segmented 3D point cloud data and motion axis estimation to enhance the precision and control of robotic actions. Experimental results suggest that our method shows promising improvements in handling large manipulation movements, indicating the potential utility of axis-based guidance in robotic manipulation tasks. 

The online axis estimation technique has demonstrated beneficial in dynamically adjusting to the real-time interaction feedback, thereby enabling the robot to maintain better control over articulated objects. 
While our method has shown enhancements over traditional approaches, 
there is still room for further refinement and optimization.

Our findings indicate that the online axis estimation process contributes positively to the accuracy and efficiency of robotic manipulation. However, this study represents an initial step, and future work is necessary to fully explore the long-term reliability and scalability of this approach across a broader spectrum of tasks and environments.

\newpage

\IEEEtriggeratref{19}

\bibliographystyle{IEEEtran}
\bibliography{IEEEexample}

\end{document}